\documentclass{article}
\usepackage{graphicx,makeidx,times}
\begin{document}


\def\ap{\textrm{'}}
\def\Biops{{\sc Biops} }
\def\Biopsn{{\sc Biops}}
\def\Oops{{\sc Oops} }
\def\Oopsn{{\sc Oops}}
\def\Aixi{{\sc Aixi} }
\def\Aixin{{\sc Aixi}}
\def\tl{{\sc Aixi}{\em (t,l)} }
\def\tln{{\sc Aixi}{\em (t,l)}}
\def\hs{{\sc Hsearch} }
\def\hsn{{\sc Hsearch}}
\def\GM{G\"{o}del Machine }
\def\gm{G\"{o}del machine }
\def\GMn{G\"{o}del Machine}
\def\gmn{G\"{o}del machine}
\newtheorem{method}{Method}[section]
\newtheorem{principle}{Principle}
\newtheorem{procedure}{Procedure}[section]
\def\odt{{\textstyle{1\over 2}}}

\title{Driven by Compression Progress: \\ A Simple Principle Explains Essential Aspects of Subjective Beauty, Novelty, Surprise, Interestingness, Attention, Curiosity, Creativity, Art, Science, Music, Jokes
}

\date{}
\author{J\"{u}rgen Schmidhuber \\
TU Munich, Boltzmannstr. 3,  85748 Garching bei M\"{u}nchen, Germany \& \\
IDSIA, Galleria 2, 6928 Manno (Lugano), Switzerland \\
{\tt juergen@idsia.ch - http://www.idsia.ch/\~{ }juergen}
}

\maketitle

\begin{abstract}
 
I argue that data becomes temporarily interesting by itself to some
self-impro\-ving, but computationally limited, subjective observer
once he learns to predict or compress the data in a better way,
thus making it subjectively simpler and more {\em beautiful.}
Curiosity is the desire to create or discover more 
non-random, non-arbitrary, regular 
data that is novel and {\em surprising} not in the 
traditional sense of Boltzmann and Shannon
but in the sense that it allows for compression progress because its 
regularity was not yet known. This drive maximizes {\em interestingness,}
the first derivative of subjective beauty or compressibility, that is, 
the steepness of the learning curve.
It motivates exploring infants, pure mathematicians, composers, artists, 
dancers, comedians, yourself, and (since 1990) artificial systems.

\end{abstract}


\vspace{0.33cm}
\noindent
{\em First version of this preprint published 23 Dec 2008; revised 15 April 2009. 
Short version:  \cite{Schmidhuber:09sice}. Long version:
 \cite{Schmidhuber:09abials}. We distill some of the essential ideas in earlier work (1990-2008) on this subject: \cite{Schmidhuber:90thesis,Schmidhuber:90diff,Schmidhuber:90sab,Schmidhuber:91cur,Schmidhuber:91singaporecur,Storck:95,Schmidhuber:97interesting,Schmidhuber:02predictable,Schmidhuber:04cur} and especially
recent papers \cite{Schmidhuber:06cs,Schmidhuber:07alt,Schmidhuber:07ds,Schmidhuber:08kes}.}

\newpage
\tableofcontents


\section{Store \& Compress \& Reward Compression Progress}
\label{basics}

If the history of the entire universe were computable \cite{Zuse:67,Zuse:69},
and there is no evidence against this possibility \cite{Schmidhuber:06random},
then its simplest explanation would be the shortest
program that computes it \cite{Schmidhuber:97brauer,Schmidhuber:00v2}.
Unfortunately there is no general way of finding
the shortest program computing any given data
\cite{Kolmogorov:65,Solomonoff:64,Solomonoff:78,LiVitanyi:97}.
Therefore physicists have traditionally proceeded incrementally, 
analyzing just a small aspect of the world at any given
time, trying to find simple laws that allow for
describing their limited observations better than the
best previously known law, essentially
trying to find a program that compresses 
the observed data better than
the best previously known program.
For example, Newton's law of gravity 
can be formulated as a short piece of code which
allows for substantially compressing many observation 
sequences involving falling apples and other objects. 
Although its predictive power is limited---for example, it does
not explain quantum fluctuations of apple atoms---it still allows 
for greatly reducing the number of bits required to encode the 
data stream, by assigning short codes to events that are 
predictable with high probability \cite{Huffman:52} under the
assumption that the law holds.  
Einstein's general relativity theory 
yields additional compression progress as it
compactly explains many previously unexplained 
deviations from Newton's predictions.

Most physicists believe
there is still room for further advances. 
Physicists, however, are not the only ones with a desire
to improve the subjective compressibility of their 
observations. Since short and simple explanations
of the past usually reflect some repetitive regularity 
that helps to predict the future as well,
{\em every} intelligent system interested in achieving
future goals should be motivated to compress the
history of raw sensory inputs in response to its actions,
simply to improve its ability to plan ahead.

A long time ago, Piaget \cite{Piaget:55} already explained the 
explorative learning behavior of children through his concepts 
of assimilation (new inputs are embedded in old schemas---this may
be viewed as a type of compression) and 
accommodation (adapting an old schema to a new input---this
may be viewed as a type of compression improvement), 
but his informal ideas did not provide enough formal details 
to permit computer implementations of his concepts.
How to model a compression progress drive in artificial systems?
Consider an active agent interacting with an initially
unknown world.  We may use our general Reinforcement Learning (RL) 
framework of artificial curiosity (1990-2008)
\cite{Schmidhuber:90thesis,Schmidhuber:90diff,Schmidhuber:90sab,Schmidhuber:91cur,Schmidhuber:91singaporecur,Storck:95,Schmidhuber:97interesting,Schmidhuber:02predictable,Schmidhuber:04cur,Schmidhuber:06cs,Schmidhuber:07ds,Schmidhuber:07alt,Schmidhuber:08kes} 
to make the agent discover data that allows for additional
compression progress and improved predictability.
The framework directs the agent towards a better understanding the world through 
active exploration, even when external reward is rare or absent, through
{\em intrinsic reward} or {\em curiosity reward} for actions leading to 
discoveries of previously unknown regularities in the action-dependent
incoming data stream. 

\subsection{Outline}
Section \ref{framework} will informally describe
our algorithmic framework based on:
(1) a continually improving predictor or compressor of the continually growing data history,
(2) a computable measure of the compressor's progress (to calculate intrinsic rewards),
(3) a reward optimizer or reinforcement learner translating rewards into action sequences expected to maximize future reward.
The formal details are left to the Appendix, which will elaborate on
the underlying theoretical concepts and describe
discrete time implementations.
Section \ref{external} will discuss the relation to external reward
(external in the sense of: originating
outside of the brain which is controlling the actions of its ``external'' body).
Section \ref{consequences} will informally
show that many
essential ingredients of intelligence
and cognition can be viewed as natural consequences of our
framework, for example, detection of 
 novelty \& surprise \& interestingness, unsupervised shifts of attention, 
subjective perception of beauty, curiosity, creativity, art, science, music, and jokes.
In particular, we reject the traditional Boltzmann / Shannon notion of surprise,
and demonstrate that both science and art can be regarded as
by-products of the desire to create / discover more data
that is compressible in hitherto unknown ways.
Section \ref{previous} will give an overview
of previous concrete implementations of approximations of our framework.
Section \ref{visual} will
apply the theory to images tailored to human observers,
illustrating the rewarding learning process 
leading from less to more subjective compressibility.
Section \ref{conclusion} will outline
how to improve our previous implementations, and
how to further test predictions of our theory in psychology and
 neuroscience.

\subsection{Algorithmic Framework}
\label{framework}

The basic ideas are embodied by the
following set of simple algorithmic principles 
distilling some of the essential ideas in previous
publications on this topic
\cite{Schmidhuber:90thesis,Schmidhuber:90diff,Schmidhuber:90sab,Schmidhuber:91cur,Schmidhuber:91singaporecur,Storck:95,Schmidhuber:97interesting,Schmidhuber:02predictable,Schmidhuber:04cur,Schmidhuber:06cs,Schmidhuber:07ds,Schmidhuber:07alt,Schmidhuber:08kes}.
As mentioned above, formal details are left to the Appendix.
As discussed in Section \ref{consequences},  
the principles at least qualitatively explain many aspects 
of intelligent agents such as humans. 
This encourages us to implement 
and evaluate them in cognitive robots and
other artificial systems. 

\begin{enumerate}
\item 
{\bf Store everything.}
During interaction with the world,
store the entire raw history of actions and 
sensory observations including reward signals---the 
data is {\em holy} as it is the only basis of 
all that can be known about the world.
To see that full data storage is not unrealistic:
A human lifetime rarely lasts much longer than $3 \times 10^{9}$ seconds.  The human
brain has roughly $10^{10}$ neurons, each with $10^4$ synapses
on average.  Assuming that only half of the brain's capacity is used 
for storing raw data, and that each synapse can store at most 6 bits,
there is still enough capacity to encode the lifelong sensory input
stream with a rate of roughly $10^5$ bits/s, comparable to the
demands of a movie with reasonable resolution. 
The storage capacity
of affordable technical systems will soon exceed this value.
If you can store the data, do not throw it away!

\item 
{\bf Improve subjective compressibility.}
In principle, any regularity in the data history 
can be used to compress it.  The compressed version of the 
data can be viewed as its simplifying explanation.
Thus, to better explain the world, 
spend some of the computation time on 
an adaptive compression algorithm trying to
partially compress the data. 
For example, an adaptive
neural network \cite{Bishop:95} may be able to learn to predict or postdict 
some of the historic data from other historic data, thus incrementally
reducing the number of bits required to encode the whole.
See Appendix \ref{performance} and \ref{improvement}.

\item
{\bf Let intrinsic curiosity reward reflect compression progress.}
The agent should monitor
the improvements of the adaptive data compressor:
whenever it learns to reduce the number of bits
required to encode the historic data, generate
an intrinsic reward signal or curiosity reward signal 
in proportion to the learning progress or compression progress, 
that is, the number of saved bits.
See Appendix \ref{improvement} and \ref{async}.

\item
{\bf Maximize intrinsic curiosity reward}
\cite{Schmidhuber:90thesis,Schmidhuber:90diff,Schmidhuber:90sab,Schmidhuber:91cur,Schmidhuber:91singaporecur,Storck:95,Schmidhuber:97interesting,Schmidhuber:02predictable,Schmidhuber:04cur,Schmidhuber:06cs,Schmidhuber:07ds,Schmidhuber:07alt}.
Let the action selector or controller
use a general Reinforcement Learning (RL) algorithm
(which should be able to observe the current state of the adaptive compressor) 
to maximize expected reward, including intrinsic curiosity reward. 
To optimize the latter, a good RL algorithm will select actions
that focus the agent's attention and learning
capabilities on those aspects of the world that
allow for finding or creating new, previously unknown but 
learnable regularities. 
In other words, it will try to maximize the steepness of the compressor's learning curve. 
This type of 
{\em active unsupervised learning} can
help to figure out how the world works. 
See Appendix \ref{optimalcur}, \ref{aixi}, \ref{gm}, \ref{rnn}.
\end{enumerate}

The framework above essentially specifies the objectives of a curious or 
creative system, not the way of achieving the objectives through the
choice of a particular adaptive compressor or predictor 
and a particular RL algorithm.
Some of the possible choices leading to special instances
of the framework (including previous concrete implementations) will be discussed later.

\subsection{Relation to External Reward}
\label{external}
Of course, the real goal of many 
cognitive systems is not just to satisfy their
curiosity, but to solve externally given problems.
Any formalizable problem can be phrased as
an RL problem for
an agent living in a possibly unknown environment,
trying to maximize the future external reward expected until
the end of its possibly finite lifetime.
The new millennium brought a few extremely general, even
universal RL algorithms 
(universal problem solvers
or universal artificial intelligences---see Appendix \ref{aixi}, \ref{gm})
that are optimal in various theoretical 
but not necessarily practical senses,
e. g., \cite{Hutter:04book+,Schmidhuber:05icann,Schmidhuber:05gmai,Schmidhuber:03newai,Schmidhuber:07newmillenniumai,Schmidhuber:06ai75,Schmidhuber:09gm}.
To the extent that learning progress / compression 
progress / curiosity as above are helpful,
these universal methods will automatically  discover and
exploit such concepts. Then why bother at all writing 
down an explicit framework for active curiosity-based experimentation?

One answer is that the present universal approaches
sweep under the carpet certain problem-independent
constant slowdowns, by burying them in the asymptotic notation of 
theoretical computer science. They
leave open an essential remaining question:
If the agent can execute only a fixed number of 
computational instructions per unit time interval
(say, 10 trillion elementary operations per second),
what is the best way of using them
to get as close as possible to the recent 
theoretical limits of universal AIs, 
especially when external rewards are very rare, 
as is the case in many realistic environments?
The premise of this paper is that 
the curiosity drive is such a general 
and generally useful concept for limited-resource
RL in rare-reward environments that it should be prewired,
as opposed to be learnt from scratch,
to save on (constant but possibly still huge) computation time.
An inherent assumption of this approach is that in
realistic worlds a better explanation of the past can only
help to better predict the future,
and to accelerate the search for
solutions to externally given tasks,
ignoring the possibility that curiosity may actually 
be harmful and ``kill the cat.''

\section{Consequences of the Compression Progress Drive}
\label{consequences}

Let us discuss how many
essential ingredients of intelligence
and cognition can be viewed as natural by-products of the
principles above.

\subsection{Compact Internal Representations or Symbols as 
By-Products of Efficient History Compression}
To compress the history of observations so far,
the compressor (say, a predictive neural network)
will automatically create internal 
representations or {\em symbols} (for example, patterns across 
certain neural feature detectors)
for things that frequently repeat themselves.  
Even when there is limited predictability,
efficient compression can still
be achieved by assigning short codes to events
that are predictable with high probability \cite{Huffman:52,SchmidhuberHeil:96}.  
For example, the sun goes up every day.
Hence it is efficient to create internal
symbols such as {\em daylight} to describe
this repetitive aspect of the data history 
by a short reusable piece of internal code, 
instead of storing just the raw data.
In fact, predictive neural networks are often 
observed to create such internal (and hierarchical) codes as a
by-product of minimizing their prediction error
on the training data.

\subsection{Consciousness as a Particular By-Product of Compression}
There is one thing that is involved in all 
actions and sensory inputs of the agent, namely, 
the agent itself. To efficiently encode the entire data 
history, it will profit from creating some sort of
internal {\em symbol} or code (e. g., a neural activity pattern)
representing the agent itself.  
Whenever this representation is actively used,
say, by activating the corresponding neurons through
new incoming sensory inputs or otherwise, the agent could
be called {\em self-aware} or {\em conscious.} 

This straight-forward explanation apparently does not 
abandon any essential aspects of our intuitive concept of consciousness,
yet seems substantially simpler than other recent views
\cite{Aleksander:05,Baars:07,Sloman:03,Seth:06,Haikonen:03,Butz:08}.
In the rest of this paper we will not have
to attach any particular mystic
value to the notion of conscious\-ness---in our view,
it is just a natural by-product of the agent's 
ongoing process of problem solving and world modeling through
data compression,
and will not play a prominent role in the remainder
of this paper.

\subsection{The Lazy Brain's Subjective, Time-Dependent Sense of Beauty}
Let $O(t)$ denote the state of some subjective observer $O$ at time $t$.
According to our {\em lazy brain theory}
\cite{Schmidhuber:97art,Schmidhuber:97femmes,Schmidhuber:98locoface,Schmidhuber:06cs,Schmidhuber:07alt,Schmidhuber:07ds},
we may identify the subjective 
beauty $B(D, O(t))$ of a new observation $D$ 
(but not its
interestingness - see Section \ref{interestingness})
as being proportional to
the number of bits required to encode $D$,
given the observer's limited
previous knowledge embodied by the
current state of its adaptive compressor.
For example, to efficiently encode previously viewed
human faces, a compressor such as a neural
network may find it useful
to generate the internal representation of 
a prototype face. To encode a new face,
it must only encode the deviations from
the prototype  \cite{Schmidhuber:97art}.
Thus a new face that
does not deviate much from the prototype
\cite{Galton:1878,Perrett:94}
will be subjectively more beautiful than others.
Similarly for faces that exhibit 
geometric regularities such as 
symmetries or simple proportions
\cite{Schmidhuber:98locoface,Schmidhuber:07ds}---in principle,
the compressor may exploit 
any regularity for reducing the number
of bits required to store the data.

Generally speaking, among several sub-patterns 
classified as {\em comparable} by a given 
observer, the subjectively most beautiful is 
the one with the simplest
(shortest) description, given the observer's current particular method
for encoding and memorizing it \cite{Schmidhuber:97art,Schmidhuber:98locoface}.
For example, mathematicians find beauty in a simple proof with a
short description in the formal language they are using.
Others like geometrically simple, aesthetically pleasing, low-complexity
drawings of various objects \cite{Schmidhuber:97art,Schmidhuber:98locoface}.

This immediately explains why many human
observers prefer faces similar to their own. 
What they see every day in the mirror will 
influence their subjective prototype face, 
for simple reasons of coding efficiency.

\subsection{Subjective Interestingness as First Derivative of Subjective \\ Beau\-ty: 
The Steepness of the Learning Curve}
\label{interestingness}
What's beautiful is not necessarily interesting.  
A beautiful thing is interesting 
only as long as it is new, that is, 
as long as the algorithmic regularity 
that makes it simple has not yet been fully assimilated by 
the adaptive observer who is still learning to 
compress the data better. It makes sense to define the
time-dependent subjective {\em Interestingness} 
$I(D, O(t))$ of data $D$ relative to observer $O$ at time $t$ by
\begin{equation}
I(D, O(t)) \sim \frac{\partial B(D, O(t))}{\partial t},
\end{equation}
the {\em first derivative} of subjective
beauty: as the learning agent improves its compression algorithm,
formerly apparently random data parts become subjectively more
regular and beautiful, requiring fewer and fewer bits
for their encoding. As long as this process is not over
the data remains interesting and rewarding.
The Appendix and Section \ref{previous} on previous implementations
will describe details of discrete time versions
of this concept.
See also \cite{Schmidhuber:91cur,Schmidhuber:91singaporecur,Storck:95,Schmidhuber:97interesting,Schmidhuber:02predictable,Schmidhuber:04cur,Schmidhuber:06cs,Schmidhuber:07ds,Schmidhuber:07alt}.

\subsection{Pristine Beauty \& Interestingness vs External Rewards}

Note that our above concepts of beauty and interestingness are limited and {\em pristine} in the sense that they are {\em not a priori} related to pleasure derived from external rewards (compare Section \ref{external}). For example, some might claim that a hot bath on a cold day triggers ``beautiful'' feelings due to rewards for achieving prewired target values of external temperature sensors (external in the sense of: outside the brain which is controlling the actions of its external body). Or a song may be called ``beautiful'' for emotional (e.g., \cite{Canamero:2003}) reasons by some who associate it with memories of external pleasure through their first kiss. Obviously this is not what we have in mind here---we are focusing solely on rewards of the intrinsic type based on learning progress.

\subsection{True Novelty \& Surprise vs Traditional Information Theory}
Consider two extreme examples of uninteresting, unsurprising,
boring data:
A vision-based agent that always stays in the dark will experience
an extremely compressible, soon totally predictable history of 
unchanging visual inputs.  
In front of a screen full of white noise 
conveying a lot of information and ``novelty'' and ``surprise''
in the traditional sense of Boltzmann and Shannon \cite{Shannon:48}, however,
it will experience highly unpredictable
and fundamentally incompressible data. 
In both cases the data is boring \cite{Schmidhuber:02predictable,Schmidhuber:07ds}
as it does not allow for further compression progress.
Therefore we reject the traditional notion of surprise.
Neither the arbitrary nor the fully predictable is {\em truly} novel or surprising---only
data with still {\em unknown} algorithmic regularities are
\cite{Schmidhuber:90thesis,Schmidhuber:90diff,Schmidhuber:90sab,Schmidhuber:91cur,Schmidhuber:91singaporecur,Storck:95,Schmidhuber:97interesting,Schmidhuber:02predictable,Schmidhuber:04cur,Schmidhuber:06cs,Schmidhuber:07ds,Schmidhuber:07alt,Schmidhuber:08kes}!

\subsection{Attention / Curiosity / Active Experimentation}
In absence of external reward,
or when there is no known way to further increase
the expected external reward,
our controller essentially tries to 
maximize {\em true novelty} or {\em interestingness,}
the {\em first derivative} of subjective beauty or compressibility, 
the steepness of the learning curve.
It will do its best to select 
action sequences expected to create observations
yielding maximal expected future compression {\em progress},  
given the limitations of both the
compressor and the compressor improvement algorithm.  
It will learn to focus its attention 
\cite{SchmidhuberHuber:91,Whitehead:92}
and its actively chosen experiments 
on things that are currently still incompressible but are
expected to become compressible / predictable
through additional learning.  It will get bored by 
things that already are subjectively compressible.
It will also get bored by things
that are currently incompressible but will apparently
remain so, given the experience so far,
or where the costs of making
them compressible exceed those of making other
things compressible, etc. \cite{Schmidhuber:90thesis,Schmidhuber:90diff,Schmidhuber:90sab,Schmidhuber:91cur,Schmidhuber:91singaporecur,Storck:95,Schmidhuber:97interesting,Schmidhuber:02predictable,Schmidhuber:04cur,Schmidhuber:06cs,Schmidhuber:07ds,Schmidhuber:07alt,Schmidhuber:08kes}.

\subsection{Discoveries}
An unusually large compression breakthrough deserves the name
{\em discovery}. For example, as mentioned in the 
introduction, the simple law of gravity 
can be described by a very short piece of code, yet it
allows for greatly compressing all previous observations of 
falling apples and other objects. 

\subsection{Beyond Standard Unsupervised Learning}

Traditional unsupervised learning is about finding regularities,
by clustering the data, or encoding it through a
factorial code \cite{Barlow:89,Schmidhuber:92ncfactorial}
with statistically independent components, or
predicting parts of it from other parts.  All of this may be viewed
as special cases of data compression. 
For example, where there are
clusters, a data point can be efficiently encoded by its cluster
center plus relatively few bits for the deviation from the center.
Where there is data redundancy, a non-redundant factorial code 
\cite{Schmidhuber:92ncfactorial} will
be more compact than the raw data.  
Where there is predictability,
compression can be achieved by assigning short codes to 
those parts of the observations
that are predictable from previous observations
with high probability \cite{Huffman:52,SchmidhuberHeil:96}.  
Generally speaking we may say
that a major goal of traditional unsupervised learning is to improve
the compression of the observed data, by discovering a program that
computes and thus explains the history (and hopefully does so quickly) but 
is clearly shorter than the shortest previously known program of this kind.

Traditional unsupervised learning is not enough though---it 
just analyzes and encodes the data but does not choose it. We have
to extend it along the dimension of active action selection, since
our unsupervised learner must also choose the actions that influence
the observed data, just like a scientist chooses his experiments,
a baby its toys, an artist his colors, a dancer his moves,
or any attentive system \cite{SchmidhuberHuber:91} its next sensory input.
That's precisely what is achieved by our
RL-based framework for curiosity and creativity.

\subsection{Art \& Music as By-Products of the Compression Progress Drive} 
\label{art}
Works of art and music may have important purposes
beyond their social aspects \cite{sciencemusic:04}  
despite of those who classify art as superfluous \cite{Pinker:97}.
Good observer-dependent 
art deepens the observer's insights about this world or
possible worlds, unveiling previously unknown regularities
in compressible data,
connecting previously disconnected patterns 
in an initially surprising way that 
makes the combination of these patterns 
subjectively more compressible (art as an eye-opener),
and eventually becomes 
known and less interesting. 
I postulate that the active
creation and attentive perception of all kinds of artwork 
are just by-products of our principle of interestingness and curiosity 
yielding reward for compressor improvements.

Let us elaborate on this idea in more detail, following the discussion 
in \cite{Schmidhuber:06cs,Schmidhuber:07ds}.
Artificial or human  
observers must perceive art sequentially, and
typically also actively, e.g., through a sequence of attention-shifting
eye saccades or camera movements scanning a sculpture, or internal
shifts of attention that filter and emphasize sounds made by a
pianist, while surpressing background noise.
Undoubtedly many derive pleasure
and rewards from perceiving works of art, such as certain paintings,
or songs.  But different subjective observers 
with different sensory apparati and compressor improvement
algorithms will prefer different input sequences.
Hence any objective theory of what is good art must 
take the subjective observer as a parameter, to 
answer questions such as:
Which sequences of actions and resulting shifts of attention should he execute to maximize
his pleasure?  According to our principle he
should select one
that maximizes the quickly learnable compressibility
that is new, relative to his current knowledge and his (usually
limited) way of incorporating / learning / compressing new data.

\subsection{Music}
For example, which song should some human observer
select next?  Not the one
he just heard ten times in a row. It became too predictable in the
process.  But also not the new weird one with the completely
unfamiliar rhythm and tonality. It seems too irregular and contain
too much arbitrariness and subjective noise.  He should try 
a song that is unfamiliar enough to contain somewhat 
unexpected harmonies or melodies or beats etc., but
familiar enough to allow for quickly recognizing the presence of a
new learnable regularity or compressibility in the sound stream. 
Sure, this song will get boring over time, but not yet.

The observer dependence is illustrated by the fact that
Sch\"{o}nberg's twelve tone music is less popular than certain 
pop music tunes, presumably because its algorithmic structure is less
obvious to many human observers as it is based on more
complicated harmonies. For example, frequency ratios of successive
notes in twelve tone music often cannot be expressed as fractions 
of very small integers. 
Those with a prior education about 
the basic concepts and
objectives and constraints of twelve tone music, however, 
tend to appreciate Sch\"{o}nberg more than those without such an
education.

All of this perfectly fits our principle: 
The learning algorithm of the compressor of a given 
subjective observer tries to better compress 
his history of acoustic and other inputs where possible. The action selector
tries to find history-influencing actions that help to improve the compressor's performance on
the history so far.  The interesting musical and other subsequences are 
those with previously unknown yet learnable types of regularities, because they lead to
compressor improvements.  The boring patterns are those that seem arbitrary
or random, or whose structure seems too hard to understand.

\subsection{Paintings, Sculpture, Dance, Film etc.}
Similar statements
not only hold for other dynamic art including
film and dance (taking into account the compressibility of controller actions), 
but also for painting and sculpture, which cause dynamic pattern sequences due to
attention-shifting actions \cite{SchmidhuberHuber:91,Whitehead:92} of the observer.

\subsection{No Objective ``Ideal Ratio'' Between Expected and Unexpected}
Some of the previous attempts at explaining aesthetic experiences in the context
of information theory \cite{Birkhoff:33,Moles:68,Bense:69,Nake:74}
emphasized the idea of an {\em ``ideal''} ratio between expected and unexpected
information conveyed by some aesthetic 
object (its {\em ``order''} vs its {\em``complexity''}). Note that
our alternative approach does
not have to postulate an objective ideal ratio of this kind. 
Instead our dynamic measure
of interestingness reflects the {\em change} in the number
of bits required to encode an object, and explicitly takes into
account  the subjective observer's prior knowledge as well as
the limitations of its compression improvement algorithm.

\subsection{Blurred Boundary Between {\em Active} Creative Artists and {\em Passive} Perceivers of Art}
\label{blurred}
Just as observers get intrinsic rewards for
sequentially focusing attention on
artwork that exhibits new, previously unknown regularities,
the {\em creative} artists get reward for making it.  
For example, I found it extremely rewarding
to discover (after hundreds of frustrating failed attempts) the simple 
geometric regularities that permitted the construction of the
drawings in Figures \ref{locoface} and \ref{butterfly}.
The distinction between artists and observers is blurred though. 
Both execute action sequences to exhibit new types of compressibility. 
The intrinsic motivations of both are 
fully compatible with our simple principle.  

Some artists, of course, crave {\em external} reward 
from other observers, in form of praise, money, or both, 
in addition to the {\em intrinsic} compression improvement-based
reward that comes from creating a truly novel work of art.
Our principle, however, conceptually separates these two reward types.

\subsection{How Artists and Scientists are Alike}
From our perspective, scientists are very much like artists.
They actively select experiments in search for simple but new
laws compressing the resulting observation history.
In particular,
the {\em creativity} of painters, dancers, musicians,
pure mathematicians, physicists, can be viewed as a 
mere by-product of our curiosity framework based on
the compression progress drive.  
All of them try to create new but non-random, non-arbitrary
data with surprising, previously unknown regularities.
For example, many physicists invent experiments to 
create data governed by previously unknown laws 
allowing to further compress the data.
On the other hand, many artists combine well-known objects in a subjectively novel way
such that the observer's subjective description of the result 
is shorter than the sum of the lengths of the descriptions of
the parts, due to some previously unnoticed
regularity shared by the parts.

What is the main difference between science and art?
The essence of science is to {\em formally nail down} the nature of 
compression progress achieved through the 
discovery of a new regularity. For example, the law of gravity can be
described by just a few symbols. In the fine arts, however,
compression progress achieved by observing
an artwork combining previously disconnected things
in a new way (art as an eye-opener) 
may be  {\em sub}conscious and not at all
formally describable by the observer, who may {\em feel} the
progress in terms of intrinsic reward without being
able to say exactly which of his memories became more
subjectively compressible in the process.

The framework in the appendix is sufficiently formal
to allow for implementation of our principle on computers. 
The resulting artificial observers will vary in terms of
the computational power of their history compressors and learning algorithms.
This will influence what is good art / science to them, and what they find
interesting.  

\subsection{Jokes and Other Sources of Fun}

Just like other entertainers and artists, comedians also tend 
to combine well-known concepts in a novel way
such that the observer's subjective description of the result 
is shorter than the sum of the lengths of the descriptions of
the parts, due to some previously unnoticed
regularity shared by the parts.

In many ways the laughs provoked by witty jokes
are similar to those provoked by the
acquisition of new skills through both babies and adults.
Past the age of 25
I learnt to juggle three balls. It was not a sudden
process but an incremental and rewarding one: in the beginning
I managed to juggle them for maybe one second 
before they fell down, then two seconds, four
seconds, etc., until I was able to do it right.  
Watching myself in the mirror (as recommended by juggling teachers)
I noticed an idiotic grin across my face whenever I made progress.
Later my little daughter grinned just
like that when she was able to stand on her own feet for the
first time.  All of this makes perfect sense within
our algorithmic framework: such grins presumably are
triggered by intrinsic reward
for generating a data stream with previously unknown
regularities, such as the sensory input sequence
corresponding to observing oneself juggling, which may be quite
different from the more familiar experience of
observing somebody else juggling, and therefore
truly novel and intrinsically rewarding, until
the adaptive predictor / compressor gets used to it.

\section{Previous Concrete Implementations of Systems Driven by 
(Approximations of) Compression Progress}
\label{previous}

As mentioned earlier, predictors and compressors are closely related.
Any type of partial predictability of the incoming sensory data stream
can be exploited to improve the compressibility of the whole.
Therefore the systems described in the first publications on 
artificial curiosity \cite{Schmidhuber:90thesis,Schmidhuber:90diff,Schmidhuber:90sab} 
already can be viewed as examples of implementations of a compression progress
drive.  

\subsection{Reward for Prediction Error (1990)}
\label{prederror}

Early work \cite{Schmidhuber:90thesis,Schmidhuber:90diff,Schmidhuber:90sab} 
described a predictor based on a recurrent neural network 
\cite{Werbos:88gasmarket,WilliamsZipser:92,RobinsonFallside:87tr,Schmidhuber:92ncn3,Pearlmutter:95,Schmidhuber:04rnn}
(in principle a rather powerful
computational device, even by today's 
machine learning standards), predicting
sensory inputs including reward signals from
the entire history of previous inputs and actions. 
The curiosity rewards were proportional to the predictor errors,
that is, it was implicitly and optimistically
assumed that the predictor will indeed improve whenever its error is high. 

\subsection{Reward for Compression Progress Through Predictor Improvements (1991)}
Follow-up work \cite{Schmidhuber:91cur,Schmidhuber:91singaporecur} 
pointed out that this approach may be inappropriate, especially
in probabilistic environments: one should not focus 
on the errors of the predictor, but on its improvements.
Otherwise the system will concentrate its search on those parts of the
environment where it can always get high prediction
errors due to noise or randomness, or due to computational
limitations of the predictor, which will prevent improvements of
the subjective compressibility of the data.
While the neural predictor of the implementation
described in the follow-up work 
was indeed computationally less powerful than the 
previous one \cite{Schmidhuber:90sab},  
there was a novelty, namely,
an explicit (neural) adaptive model of the predictor's improvements.
This model essentially learned to predict the predictor's changes.
For example, although noise was unpredictable and led to
wildly varying target signals for the predictor, in the long run these
signals did not change the adaptive predictor parameters much, and the predictor
of predictor changes was able to learn this.
A standard RL algorithm \cite{Watkins:89,Kaelbling:96,Sutton:98}
was fed with curiosity reward signals proportional to 
the expected long-term predictor changes, and thus tried
to maximize information gain 
\cite{Fedorov:72,Hwang:91,MacKay:92c,Plutowski:93,Cohn:94}
within the given limitations. In fact, we may say that
the system tried to maximize an approximation of the (discounted)
sum of the expected
first derivatives of the data's subjective predictability,
thus also maximizing an approximation of the (discounted) sum
of the expected
changes of the data's subjective compressibility.

\subsection{Reward for Relative Entropy between Agent's Prior and Posterior (1995)}
\label{entropy}

Additional follow-up work yielded an information 
theory-oriented variant of the approach in
non-\-deterministic worlds \cite{Storck:95} (1995).
The curiosity reward was again proportional to 
the predictor's surprise / information gain, this time measured 
as the Kullback-Leibler distance \cite{Kullback:59} between the learning predictor's 
subjective probability distributions before and after new 
observations - the relative entropy between its prior and posterior. 

In 2005 Baldi and Itti called this approach ``Bayesian surprise''
and demonstrated experimentally that it explains certain patterns of
human visual attention 
better than certain previous approaches \cite{itti:05}.

Note that the concepts of Huffman coding \cite{Huffman:52} and
relative entropy between prior and posterior 
immediately translate into a
measure of learning progress reflecting
the number of saved bits---a measure of improved data compression.

Note also, however, that the naive probabilistic approach to
data compression is unable to
discover more general types of  {\em algorithmic} compressibility
\cite{Solomonoff:64,Kolmogorov:65,LiVitanyi:97,Schmidhuber:02ijfcs}.
For example, the decimal expansion of $\pi$ looks random and incompressible
but isn't:  there is a very short algorithm computing all of  $\pi$,  yet
any finite sequence of digits will occur in $\pi$'s  expansion
as frequently as expected if $\pi$ were truly random, 
that is, no simple statistical learner will 
outperform random guessing at predicting the next
digit from a limited time window of previous digits.
More general {\em program} search techniques 
(e.g., \cite{Levin:73,Schmidhuber:04oops,Cramer:85,Olsson:95})
are necessary to extract the underlying algorithmic regularity.

\subsection{Zero Sum Reward Games for Compression 
Progress Revealed by Algorithmic Experiments (1997)}

More recent work \cite{Schmidhuber:97interesting,Schmidhuber:02predictable} (1997)
greatly increased the computational power of controller and predictor by
implementing them as co-evolving, symmetric, opposing modules consisting of self-modifying
probabilistic programs 
\cite{Schmidhuber:97ssa,Schmidhuber:97bias}
written in a universal programming language
\cite{Goedel:31,Turing:36} allowing for loops, recursion,
and hierarchical structures.
The internal storage for temporary computational results
of the programs was viewed as part of the changing environment.
Each module could suggest experiments 
in the form of probabilistic algorithms to be executed,
and make confident predictions about their effects 
by betting on their outcomes, where the {\em `betting money'}
essentially played the role of the intrinsic reward.
The opposing module could reject or accept 
the bet in a zero-sum game by making a 
contrary prediction. In case of acceptance,
the winner was determined by executing the algorithmic experiment and 
checking its outcome;
the money was eventually transferred from the surprised loser to the 
confirmed winner.  Both modules tried to maximize their money using
a rather general RL algorithm designed for 
complex stochastic policies \cite{Schmidhuber:97ssa,Schmidhuber:97bias} (alternative
RL algorithms could be plugged in as well).
Thus both modules were motivated to discover {\em truly novel} algorithmic
regularity / compressibility, where the subjective baseline
for novelty was given by what the opponent already knew about
the world's repetitive regularities.

The method can be viewed as
system identification through co-evolution of computable models and tests.
In 2005 a similar co-evolutionary approach based on less general
models and tests was implemented
by Bongard and Lipson \cite{Bongard:05}.

\subsection{Improving Real Reward Intake}

Our references above demonstrated experimentally
that the presence of intrinsic reward or curiosity reward 
actually can speed up the collection of {\em external} reward.

\subsection{Other Implementations}

Recently several researchers also implemented variants 
or approximations of the curiosity framework.
Singh and Barto and coworkers focused on implementations 
within the option framework of RL \cite{Barto:04,Singh:05nips},
directly using prediction errors as curiosity rewards 
as in Section \ref{prederror}
\cite{Schmidhuber:90thesis,Schmidhuber:90diff,Schmidhuber:90sab} 
---they actually were the ones who 
coined the expressions {\em intrinsic reward}
and {\em intrinsically motivated} RL.
Additional implementations were presented at the
2005 AAAI Spring Symposium on Developmental Robotics 
\cite{Blank:05ws};
compare the Connection Science Special Issue \cite{BM06}.

\section{Visual Illustrations of Subjective Beauty and 
its {\em First Derivative} Interestingness}
\label{visual}

As mentioned above (Section \ref{entropy}), the probabilistic
variant of our theory \cite{Storck:95} (1995) was able
to explain certain shifts of human visual attention \cite{itti:05} (2005).
But we can also apply our approach to the complementary problem of {\em constructing}
images that contain quickly learnable regularities, arguing again that there 
is no fundamental difference between the motivation of creative artists 
and passive observers of visual art (Section \ref{blurred}).
Both create action sequences yielding
interesting inputs, where interestingness is a  measure of learning progress,
for example, based on the relative entropy between 
prior and posterior (Section \ref{entropy}), or the saved 
number of bits needed to encode the data (Section \ref{basics}), or 
something similar (Section \ref{previous}).

Here we provide 
examples of subjective beauty tailored to human observers,
and illustrate the learning process 
leading from less to more subjective beauty. Due to the nature
of the present written medium, we have to use visual examples 
instead of acoustic or tactile ones.
Our examples are intended to support the hypothesis that
unsupervised {\em attention} and
the {\em creativity} of artists, dancers, musicians,
pure mathematicians 
are just by-products of their compression progress drives.

\subsection{A Pretty Simple Face with a  Short Algorithmic Description}

\begin{figure}[hbt]
\centering
\includegraphics[height=11.3cm]{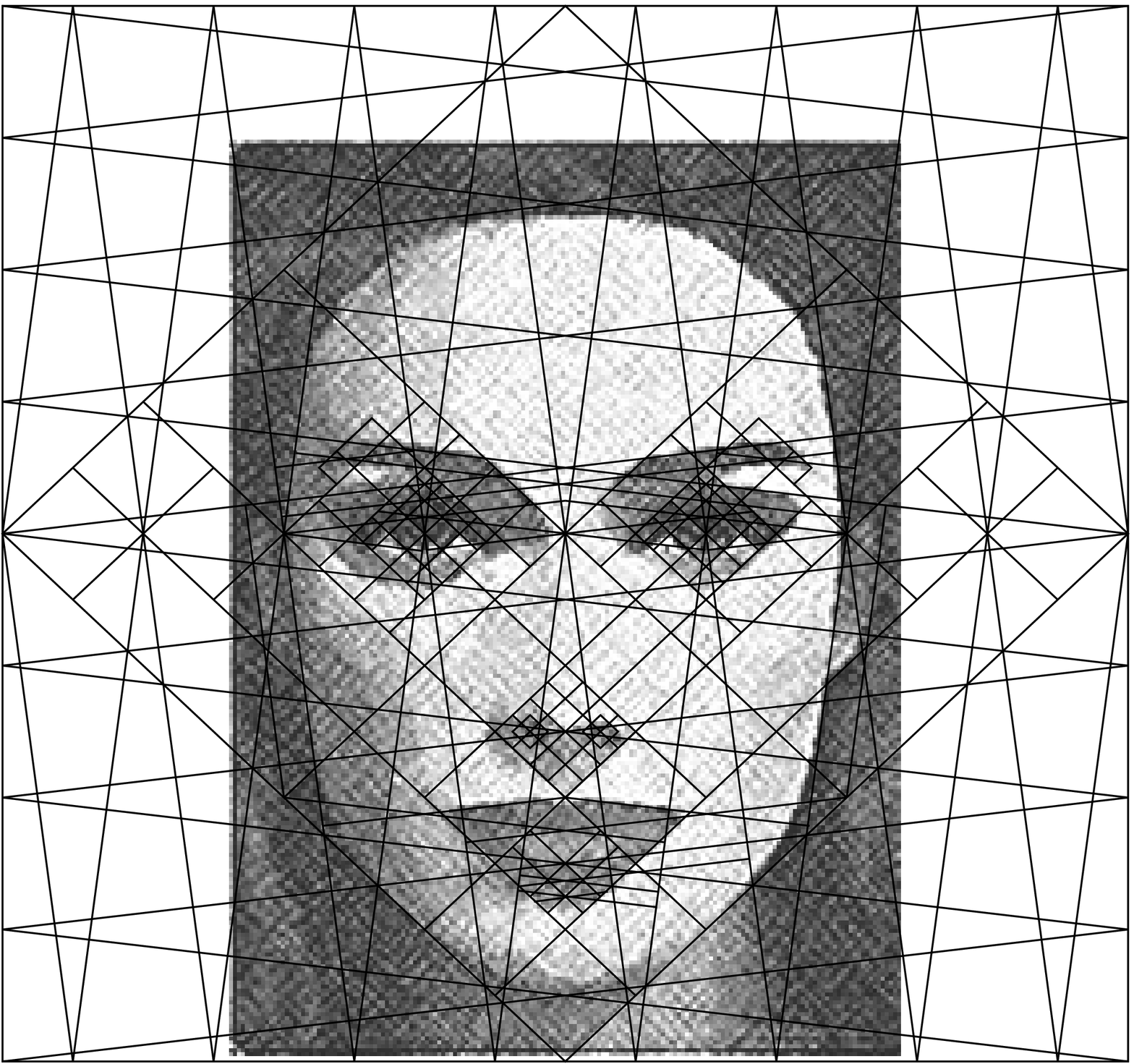}
\caption{Previously published construction plan 
\cite{Schmidhuber:98locoface,Schmidhuber:07ds}
of a female face (1998).
Some human observers report they feel this face is `beautiful.'
Although the drawing has lots of noisy details (texture etc) without an 
obvious short description, positions and shapes of the basic facial
features are compactly encodable through a very simple geometrical scheme, 
simpler and much more precise than ancient facial proportion 
studies by Leonardo da Vinci
and Albrecht D\"{u}rer.
Hence the image contains a highly compressible algorithmic 
regularity or pattern 
describable by few bits of information.
An observer can perceive it through a sequence 
of attentive eye movements or saccades, and consciously 
or subconsciously discover 
the compressibility of the incoming data stream.
How was the picture made?
First the sides of a square were partitioned into $2^4$ equal intervals.
Certain interval boundaries were connected to obtain
three rotated, superimposed grids based on lines with slopes
$\pm 1$ or $\pm 1/2^3$ or $\pm 2^3/1$.
Higher-resolution details of the grids were obtained 
by iteratively selecting
two previously generated, neighboring, parallel lines and inserting
a new one equidistant to both.
Finally the grids were vertically compressed by a factor of $1-2^{-4}$.
The resulting lines and their intersections define
essential boundaries and shapes of eyebrows, eyes, lid shades,
mouth, nose, and facial frame in a simple way that 
is obvious from the construction plan.
Although this plan is simple in hindsight,
it was hard to find: hundreds of my previous attempts at discovering
such precise matches between simple geometries and pretty faces failed.}
\label{locoface}
\end{figure}

Figure \ref{locoface} depicts the construction
plan of a female face considered
{\em `beautiful'} by some human observers. It also shows that the 
essential features of this face follow a very simple 
geometrical pattern \cite{Schmidhuber:98locoface} that can be
specified by very few bits of information. That is,
the data stream generated by observing the image (say, through
a sequence of eye saccades) is more compressible than it would
be in the absence of such regularities. 
Although few people are able to immediately see 
how the drawing was made in absence of its superimposed grid-based
explanation, most 
do notice that the facial features somehow fit together and
exhibit some sort of regularity. According to our postulate,
the observer's reward is generated by the conscious or 
subconscious discovery of this compressibility.
The face remains interesting
until its observation does not reveal any additional
previously unknown regularities. Then it becomes
boring even in the eyes of those who 
think it is beautiful---as has been pointed out repeatedly above,
beauty and interestingness are two different things.

\subsection{Another Drawing That Can Be Encoded By Very Few Bits}

\begin{figure}[hbt]
\centering
\includegraphics[height=11.3cm]{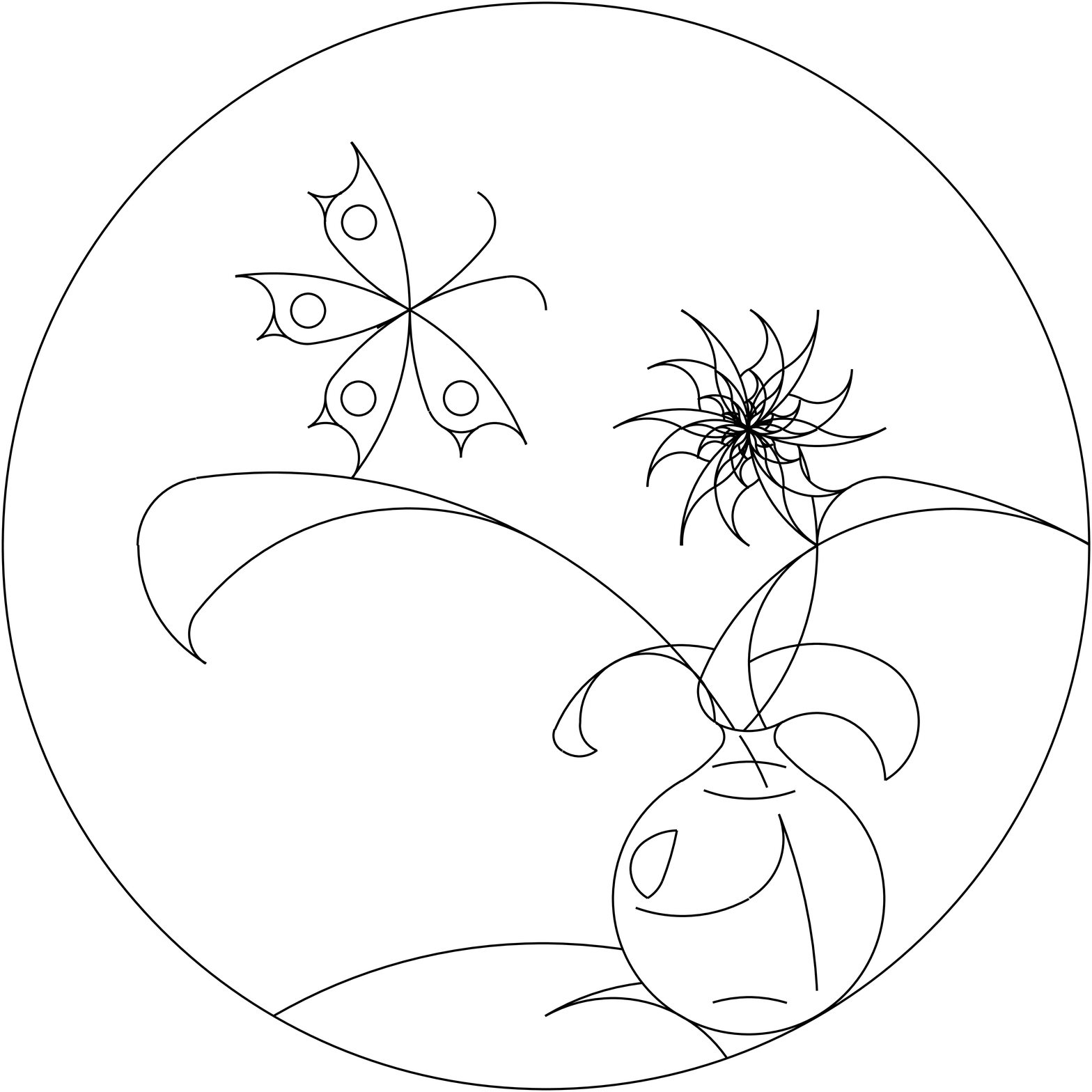}
\caption{
Image of a butterfly and a vase with a flower,
reprinted from {\em Leonardo}
\cite{Schmidhuber:97art,Schmidhuber:06cs}.
An explanation of how the image was constructed
and why it has a very short description
is given in Figure \ref{butterflyex}. 
}
\label{butterfly}
\end{figure}

\begin{figure}[hbt]
\centering
\includegraphics[height=11.3cm]{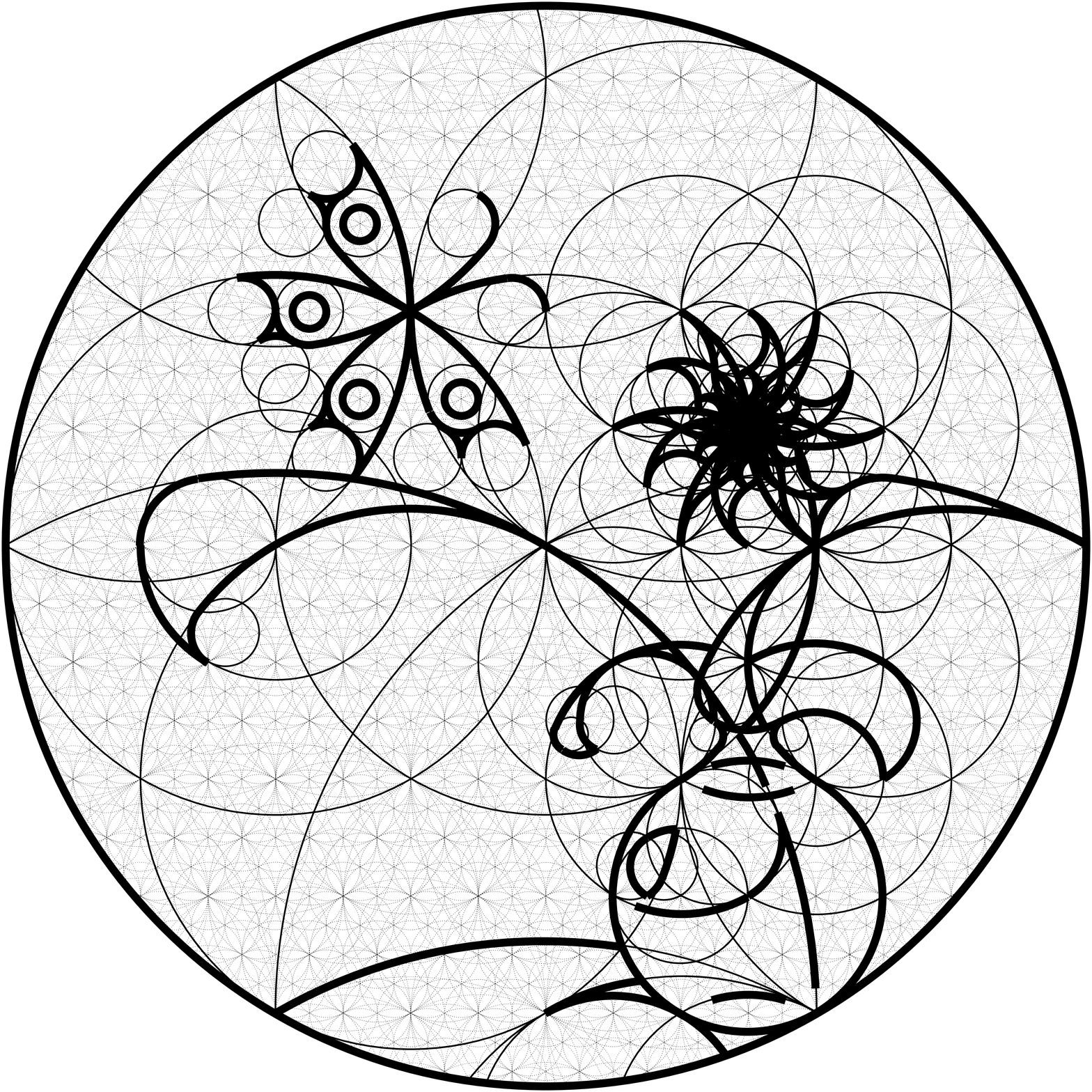}
\caption{
Explanation of how Figure \ref{butterfly} was constructed
through a very simple algorithm exploiting fractal circles
\cite{Schmidhuber:97art}.  The frame is a circle; its leftmost
point is the center of another circle of the same size.
Wherever two circles of equal size touch or intersect
are centers of two more circles with equal and half size, respectively.
Each line of the drawing is a segment of some circle, its
endpoints are where circles touch or intersect.
There are few big circles and many small ones. In general,
the smaller a circle, the more bits are needed to specify it.
The drawing is simple (compressible) as it is based on few, rather large circles.
Many human observers report that they derive a certain amount of pleasure from
discovering this simplicity. The observer's learning process causes
a reduction of the subjective complexity of the data,
yielding a temporarily high derivative of subjective beauty:
a temporarily steep learning curve.
(Again I needed a long time to discover
a satisfactory and rewarding way of using fractal circles to
create a reasonable drawing.)
}
\label{butterflyex}
\end{figure}

Figure \ref{butterfly}
provides another example: a butterfly and a vase with a flower.  
It can be specified by very few bits of
information as it can be constructed through a very simple procedure
or algorithm based on fractal circle patterns  
\cite{Schmidhuber:97art}---see Figure \ref{butterflyex}.
People who understand this algorithm tend to appreciate
the drawing more than those who do not. 
They realize how simple it is. 
This is not an immediate,
all-or-nothing, binary process though. 
Since the typical human visual system has a lot of
experience with circles, 
most people quickly notice that the curves somehow 
fit together in a regular way. But few are
able to immediately state the precise 
geometric principles underlying the drawing \cite{Schmidhuber:06cs}.
This pattern, however, is learnable from 
Figure \ref{butterflyex}.
The conscious or subconscious discovery process 
leading from a longer to a shorter
description of the data, or from less to more compression,
or from less to more subjectively perceived beauty, yields
reward depending on the first derivative of subjective beauty, that is,
the steepness of the learning curve.

\section{Conclusion \& Outlook}
\label{conclusion}

We pointed out that a surprisingly simple algorithmic principle based on the
notions of data compression and data compression {\em progress} informally explains fundamental aspects
of attention, novelty, surprise, interestingness, curiosity, creativity, subjective beauty, jokes,
and science \& art in general.  The crucial ingredients of the corresponding {\em formal} framework are 
(1) a continually improving predictor or compressor of the continually growing data history,
(2) a computable measure of the compressor's progress (to calculate intrinsic rewards),
(3) a reward optimizer or reinforcement learner translating rewards into action sequences expected to maximize future reward.
To improve our previous implementations of these ingredients (Section \ref{previous}), we will  
(1) study better adaptive compressors, in particular,
recent, novel RNNs \cite{Schmidhuber:09rnnbook} and other general but
 practically feasible methods for
making predictions \cite{Schmidhuber:04oops}; 
(2) investigate under which conditions learning progress measures can be computed both accurately and efficiently, 
without  frequent expensive compressor performance evaluations on the entire history so far;  
(3) study the applicability of recent improved RL 
techniques in the fields of policy gradients \cite{Sutton:99,wierstraECML07,wierstraCEC08,rueckstiess2008b,sehnke2008b,wierstraPPSN08},
artificial evolution \cite{moriarty:ml96,gomez:ab97,gomez:ijcai99,gomez:phd,Gomez:03+,Gomez:05geccocern,Gomez:08jmlr},
and others \cite{Schmidhuber:01direct,Schmidhuber:04oops}.

Apart from building improved {\em artificial} curious agents,
we can test the predictions of our theory in psychological investigations of {\em human} behavior,
extending previous studies in this vein \cite{itti:05} and going beyond
anecdotal evidence mentioned above.
It should be easy to devise controlled experiments where test subjects must
anticipate initially unknown but causally connected event sequences exhibiting more or less 
complex, learnable patterns or regularities. The subjects will be asked
to quantify their intrinsic rewards in response to their improved predictions. Is the reward indeed
strongest when the predictions are improving most rapidly? Does the intrinsic reward indeed vanish
as the predictions become perfect or do not improve any more?

Finally, how to test our predictions through studies in neuroscience?
Currently we hardly understand the human neural machinery.
But it is well-known that certain neurons seem to predict others,
and brain scans show how certain brain areas light up in response to reward.
Therefore the psychological experiments suggested above should be accompanied by
neurophysiological studies to localize the origins of intrinsic
rewards, possibly linking them to improvements of
neural predictors. 

Success in this endeavor would provide additional motivation
to implement our principle on robots.

\appendix

\section{Appendix}

This appendix is based in part on references \cite{Schmidhuber:06cs,Schmidhuber:07ds}.

The world can be explained to a degree by compressing it.
Discoveries correspond to large data compression improvements 
(found by the given, application-dependent compressor improvement algorithm).
How to build an adaptive agent that not only tries to
achieve externally given rewards but also 
to discover, in an unsupervised and experiment-based fashion, 
explainable and compressible data?
(The explanations gained through explorative behavior may 
eventually help to solve teacher-given tasks.)

Let us formally consider a learning agent whose single life 
consists of discrete cycles or time steps $t=1, 2, \ldots, T$.
Its complete lifetime $T$ may or may not be known in advance.
In what follows, the value of any time-varying variable $Q$
at time $t$ ($1 \leq  t \leq T$) will be denoted by $Q(t)$,
the ordered sequence of values $Q(1),\ldots,Q(t)$ by $Q(\leq \! t)$,
and the (possibly empty) sequence $Q(1),\ldots,Q(t-1)$ by $Q(< t)$.
At any given $t$  the agent receives a real-valued input $x(t)$ from
the environment and executes a real-valued 
action $y(t)$ which may affect future inputs. At times $t<T$ its goal
is to maximize future success or {\em utility}
\begin{equation}
\label{u}
u(t) =
E_{\mu} \left [ \sum_{\tau=t+1}^T  
r(\tau)~~ \Bigg| ~~ h(\leq \! t) \right ],
\end{equation}
where $r(t)$ is an additional real-valued reward input at time $t$,
$h(t)$ the ordered triple $[x(t), y(t), r(t)]$
(hence $h(\leq \! t)$ is the known history up to $t$),
and $E_{\mu}(\cdot \mid \cdot)$ denotes the conditional expectation operator
with respect to some possibly unknown distribution $\mu$ from a set $\cal M$
of possible distributions. Here $\cal M$ reflects
whatever is known about the possibly probabilistic reactions
of the environment.  For example, $\cal M$ may contain all computable
distributions \cite{Solomonoff:64,Solomonoff:78,LiVitanyi:97,Hutter:04book+}.
There is just one life, no need for predefined repeatable trials, 
no restriction to Markovian 
interfaces between sensors and environment,
and the utility function implicitly takes into account the 
expected remaining lifespan $E_{\mu}(T \mid  h(\leq \! t))$
and thus the possibility to extend it through appropriate actions
\cite{Schmidhuber:05icann,Schmidhuber:05gmai,Schmidhuber:05gmconscious,Schmidhuber:09gm}.

Recent work has led to the first learning machines
that are universal and optimal in various very general senses 
\cite{Hutter:04book+,Schmidhuber:05icann,Schmidhuber:05gmai}.
As mentioned in the introduction,
such machines can in principle find out by themselves whether
curiosity and
world model construction are useful or useless in a given
environment, and learn to behave accordingly.  
The present appendix, however, will assume {\em a priori} that
compression / explanation of the history is good and should be done;
here we shall not worry about the possibility
that curiosity can be harmful and ``kill the cat.''
Towards this end, in the spirit of our previous work since 1990
\cite{Schmidhuber:90thesis,Schmidhuber:90diff,Schmidhuber:90sab,Schmidhuber:91cur,Schmidhuber:91singaporecur,Storck:95,Schmidhuber:97interesting,Schmidhuber:02predictable,Schmidhuber:04cur,Schmidhuber:06cs,Schmidhuber:07ds,Schmidhuber:07alt,Schmidhuber:08kes} 
we split the reward signal $r(t)$ into
two scalar real-valued components: $r(t)=g(r_{ext}(t),r_{int}(t))$,
where $g$ maps pairs of real values to real values,  e.g., $g(a,b)=a+b$. 
Here $r_{ext}(t)$ denotes traditional {\em external} reward provided
by the environment, such as negative reward in 
response to bumping against a wall, or positive
reward in response to reaching some teacher-given goal state. 
But for the purposes of this paper we are especially
interested in $r_{int}(t)$, the internal or intrinsic 
or {\em curiosity} reward, which is provided whenever
the data compressor / internal world model of the agent improves in
some measurable sense. Our initial focus will be on
the case $r_{ext}(t)=0$ for all valid $t$.
The basic principle is essentially the one we published before in various variants
\cite{Schmidhuber:90thesis,Schmidhuber:90diff,Schmidhuber:90sab,Schmidhuber:91cur,Schmidhuber:91singaporecur,Storck:95,Schmidhuber:97interesting,Schmidhuber:02predictable,Schmidhuber:04cur,Schmidhuber:06cs,Schmidhuber:07ds,Schmidhuber:07alt}:
\begin{principle}
\label{principle}
Generate curiosity reward for the controller in response to improvements of the predictor or history compressor.
\end{principle}
So we conceptually separate the goal (explaining / compressing the history)
from the means of achieving the goal. Once the goal is
formally specified in terms of an algorithm for computing curiosity rewards, 
let the controller's reinforcement learning (RL) mechanism figure 
out how to translate such rewards
into action sequences that allow the given compressor improvement algorithm
to find and exploit previously unknown types of compressibility.

\subsection{Predictors vs Compressors}

Much of our previous work on artificial curiosity
was prediction-oriented, e. g.,
\cite{Schmidhuber:90thesis,Schmidhuber:90diff,Schmidhuber:90sab,Schmidhuber:91cur,Schmidhuber:91singaporecur,Storck:95,Schmidhuber:97interesting,Schmidhuber:02predictable,Schmidhuber:04cur}.
Prediction and compression are closely related though.
A predictor that correctly predicts 
many $x(\tau)$, given history $h(< \tau)$, for $1 \leq \tau \leq  t$,  
can be used to encode $h(\leq \! t)$ compactly.
Given the predictor, only the wrongly predicted $x(\tau)$ plus
information about the
corresponding time steps $\tau$ are necessary
to reconstruct history $h(\leq \! t)$, e.g.,  \cite{Schmidhuber:92ncchunker}.
Similarly, a predictor that learns a probability distribution of
the possible next events, given previous events,  can be used to
efficiently encode observations with high (respectively low) predicted probability 
by few (respectively many) bits \cite{Huffman:52,SchmidhuberHeil:96}, thus achieving
a compressed history representation.
Generally speaking, we may view the predictor as the essential part of a 
program $p$ that re-computes $h(\leq \! t)$. 
If this program is short in comparison to the raw data $h(\leq \! t)$, then 
$h(\leq \! t)$ is regular or non-random 
\cite{Solomonoff:64,Kolmogorov:65,LiVitanyi:97,Schmidhuber:02ijfcs},
presumably reflecting essential environmental laws. Then
$p$ may also be highly useful for predicting future, yet unseen
 $x(\tau)$ for $\tau>t$.  

It should be mentioned, however, that the compressor-oriented approach
to prediction based on the principle of Minimum Description Length (MDL)
\cite{Kolmogorov:65,Wallace:68,Wallace:87,Rissanen:78,LiVitanyi:97}
does not necessarily converge to the correct predictions
as quickly as Solomonoff's universal inductive inference  
\cite{Solomonoff:64,Solomonoff:78,LiVitanyi:97}, although both approaches
converge in the limit under general conditions \cite{Poland:05mdlreg}.

\subsection{Which Predictor or History Compressor?}
\label{predictor}
The complexity of evaluating some compressor $p$ on
history
$h(\leq t)$ depends on both $p$ and its performance
measure $C$. Let us first focus on the former.
Given $t$, one of the simplest $p$ will 
just use a linear mapping to
predict $x(t+1)$ from $x(t)$ and $y(t+1)$. 
More complex $p$ such as adaptive recurrent neural networks (RNN)
\cite{Werbos:88gasmarket,WilliamsZipser:92,RobinsonFallside:87tr,Schmidhuber:92ncn3,Pearlmutter:95,Hochreiter:97lstm,Schmidhuber:03rnnaissance,Schmidhuber:04learningrobots,Schmidhuber:04rnn}
will use a nonlinear mapping
and possibly the entire history $h(\leq t)$ as a basis for the predictions.
In fact, the first work on artificial curiosity \cite{Schmidhuber:90sab} focused
on online learning RNN of this type. 
A theoretically optimal predictor would be Solomonoff's
above-mentioned universal induction scheme
\cite{Solomonoff:64,Solomonoff:78,LiVitanyi:97}.

\subsection{Compressor Performance Measures}
\label{performance}
At any time $t$ ($1 \leq  t < T$), 
given some compressor program $p$ able to compress
history $h(\leq \! t)$, let $C(p,h(\leq \! t))$ denote 
$p$'s compression performance on $h(\leq \! t)$. 
An appropriate performance measure would be
\begin{equation}
C_l(p,h(\leq \! t))=l(p),
\end{equation}
where $l(p)$ denotes the
length of $p$, measured in number of bits: the shorter $p$,
the more algorithmic regularity and compressibility and 
predictability and lawfulness in the observations so far.
The ultimate limit for $C_l(p,h(\leq \! t))$ would be 
$K^*(h(\leq \! t))$, a variant of the Kolmogorov complexity
of $h(\leq \! t)$, namely, the length of the shortest program 
(for the given hardware) that computes an output
starting with $h(\leq \! t)$ 
\cite{Solomonoff:64,Kolmogorov:65,LiVitanyi:97,Schmidhuber:02ijfcs}.

\subsection{Compressor Performance Measures Taking Time Into Account}
$C_l(p,h(\leq \! t))$ does not take into account the time
$\tau(p,h(\leq \! t))$ spent by $p$ on computing $h(\leq \! t)$.
An alternative performance measure inspired by concepts
of optimal universal search \cite{Levin:73,Schmidhuber:04oops} is
\begin{equation}
C_{l \tau}(p,h(\leq \! t))= l(p) + \log~\tau(p,h(\leq \! t)).
\end{equation}
Here compression by one bit is worth as much as runtime
reduction by a factor of $\frac{1}{2}$.
From an asymptotic optimality-oriented point of view
this is one of the best ways of trading off 
storage and computation time
\cite{Levin:73,Schmidhuber:04oops}.

\subsection{Measures of Compressor Progress / Learning Progress}
\label{improvement}
The previous sections  only discussed measures
of compressor performance, but not of performance {\em improvement,}
which is the essential issue in our curiosity-oriented context.
To repeat the point made above:
 {\em The important thing are the improvements 
of the compressor, not its compression performance per se.}
Our curiosity reward in response to the 
compressor's progress 
(due to some application-dependent compressor improvement algorithm)
between times $t$ and $t+1$ should be
\begin{equation}
r_{int}(t+1)= f[C(p(t),h(\leq \! t+1)),C(p(t+1),h(\leq \! t+1))],
\end{equation}
where $f$ maps pairs of
real values to real values.  Various alternative progress measures
are possible; most obvious is $f(a,b)=a-b$. This corresponds to a 
discrete time version of
maximizing the first derivative of subjective data compressibility.

{\em Note that both the old and the new compressor have to be tested on the
same data, namely, the history so far.}

\subsection{Asynchronous Framework for Creating Curiosity Reward}
\label{async}

Let $p(t)$ denote the agent's current compressor program at time $t$,
$s(t)$ its current controller, and do: 

\noindent \\
{\bf Controller:}
At any time $t$ ($1 \leq  t < T$) do:
\begin{enumerate}
\item
Let $s(t)$ use (parts of) history $h(\leq  t)$
to select and execute $y(t+1)$. 
\item
Observe $x(t+1)$.
\item
Check if there is non-zero curiosity reward $r_{int}(t+1)$
provided by the separate, asynchronously running
compressor improvement algorithm (see below).
If not, set $r_{int}(t+1)=0$.
\item
Let the controller's reinforcement learning (RL) algorithm use $h(\leq \! t+1)$
including $r_{int}(t+1)$
(and possibly also the latest available 
compressed version of the observed data---see below)
to obtain a new controller $s(t+1)$, 
in line with objective (\ref{u}).
\end{enumerate}

\noindent
{\bf Compressor:}
Set $p_{new}$ equal to the initial data compressor.
Starting at time 1, repeat forever until interrupted by death at time $T$:
\begin{enumerate}
\item
Set $p_{old}=p_{new}$; 
get current time step $t$ and set
$h_{old}=h(\leq \! t)$. 
\item
Evaluate $p_{old}$ on $h_{old}$, to obtain $C(p_{old},h_{old})$
(Section \ref{performance}).
This may take many time steps.
\item
Let some (application-dependent)
compressor improvement algorithm 
(such as a learning algorithm for
an adaptive neural network predictor)
use $h_{old}$
to obtain a hopefully better compressor $p_{new}$ 
(such as a neural net with the same size but
improved prediction capability
and therefore improved compression performance
\cite{SchmidhuberHeil:96}). 
Although this may take many time steps (and could
be partially performed during ``sleep''), $p_{new}$ 
may not be optimal, due to limitations of
the learning algorithm, e.g., local maxima.
\item
Evaluate $p_{new}$ on $h_{old}$, to obtain $C(p_{new},h_{old})$.
This may take many time steps.
\item
\label{fasync}
Get current time step $\tau$ and  generate curiosity reward 
\begin{equation}
r_{int}(\tau)=f[C(p_{old},h_{old}),C(p_{new},h_{old})], 
\end{equation}
e.g., $f(a,b)=a-b$; see Section \ref{improvement}.
\end{enumerate}
Obviously this asynchronuous scheme
may cause long temporal delays 
between controller actions and corresponding 
curiosity rewards. This may impose a heavy
burden on the controller's RL algorithm whose task
is to assign credit to past actions
(to inform the controller about beginnings of compressor 
evaluation processes etc., 
we may augment its input by unique representations of such events).
Nevertheless, there are 
RL algorithms for this purpose which are
theoretically optimal in various senses,
to be discussed next.

\subsection{Optimal Curiosity \& Creativity \& Focus of Attention}
\label{optimalcur}

Our chosen compressor class typically will have
certain computational limitations.  In the absence of any external rewards,
we may define {\em optimal pure curiosity
behavior} relative to these limitations:
At time $t$ this behavior would select the action that maximizes
\begin{equation}
\label{optcur}
u(t) = E_{\mu} \left [ \sum_{\tau=t+1}^{T} 
r_{int}(\tau)~~ \Bigg| ~~ h(\leq \! t) \right ].
\end{equation}
Since the true, world-governing
probability distribution $\mu$ is unknown, 
the resulting task of the controller's RL algorithm 
may be a formidable one. 
As the system is revisiting previously incompressible parts of the environment,
some of those will tend to  become more subjectively
compressible, and the corresponding
curiosity rewards will
decrease over time. A good RL algorithm must somehow
detect and then {\em predict} this decrease, and act accordingly.  
Traditional RL algorithms \cite{Kaelbling:96}, however,
do not provide any theoretical guarantee of optimality for such
situations. 
(This is not to say though that sub-optimal
RL methods may not lead to success in certain applications; 
experimental studies might lead to interesting insights.)

Let us first make the natural assumption that the compressor 
is not super-complex such as Kolmogorov's, that is, its output 
and $r_{int}(t)$ are computable for all $t$. 
Is there a best possible RL algorithm that comes as
close as any other to maximizing
objective (\ref{optcur})? Indeed, there is. 
Its drawback, however, is that it is not
computable in finite time. Nevertheless,
it serves as a reference point for defining
what is achievable at best.

\subsection{Optimal But Incomputable Action Selector}
\label{aixi}

There is an optimal way of selecting actions
which makes use of Solomonoff's theoretically optimal
universal predictors and their Bayesian learning algorithms
\cite{Solomonoff:64,Solomonoff:78,LiVitanyi:97,Hutter:04book+,Hutter:07uspx}.
The latter only assume that the reactions of the environment are sampled from 
an unknown probability distribution $\mu$ contained in a set $\cal M$
of all enumerable distributions---compare text after equation (\ref{u}).
More precisely, given an observation sequence $q(\leq \! t)$ 
we want to use the Bayes formula to predict the probability 
of the next possible $q(t+1)$.
Our only assumption is that there exists a computer program
that can take any $q(\leq \! t)$ as an input and compute its {\em a priori} probability
according to the $\mu$ prior.
In general we do not know this program, hence
we predict using  a mixture prior instead:
\begin{equation}
\label{xi}
\xi(q(\leq \! t)) =\sum_i w_i\mu_i (q(\leq \! t)),
\end{equation}
a weighted sum of {\em all} distributions $\mu_i \in \cal M$, $i=1, 2, \ldots$, 
where the sum of the constant positive weights satisfies $\sum_i w_i \leq 1$. 
This is indeed the best one can possibly do, in a very general sense
\cite{Solomonoff:78,Hutter:04book+}. The drawback
of the scheme is its incomputability, since $\cal M$ contains
infinitely many distributions.
We may increase the theoretical power of the scheme by
augmenting $\cal M$ by certain
non-enumerable but
limit-computable distributions \cite{Schmidhuber:02ijfcs},
or restrict it such that it becomes computable,
e.g., by assuming the world is computed
by some unknown but deterministic computer
program sampled from the Speed Prior \cite{Schmidhuber:02colt} which assigns
low probability to environments that are hard to compute by any method.

Once we have such an optimal predictor, we can extend it
by formally including the effects of executed actions to define an 
optimal action selector maximizing future expected reward.
At any time $t$, Hutter's theoretically optimal
(yet uncomputable) RL algorithm \Aixi \cite{Hutter:04book+} 
uses an extended version of Solomonoff's prediction scheme  
to select those action sequences that promise maximal
future reward up to some horizon $T$,  given the current data $h(\leq \! t)$.
That is, in cycle $t+1$, \Aixi
selects as its next action the first action of an action sequence
maximizing $\xi$-predicted reward up to the given horizon, appropriately
generalizing eq. (\ref{xi}).
\Aixi uses observations optimally \cite{Hutter:04book+}: 
the Bayes-optimal policy $p^\xi$ based on
the mixture $\xi$ is self-optimizing in the sense that its average
utility value converges asymptotically for all $\mu \in \cal M$ to the
optimal value achieved by the Bayes-optimal policy $p^\mu$
which knows $\mu$ in advance. The necessary and sufficient condition is
that $\cal M$ admits self-optimizing policies.
The policy $p^\xi$ is also Pareto-optimal
in the sense that there is no other policy yielding higher or equal
value in {\em all} environments $\nu \in \cal M$ and a strictly higher
value in at least one \cite{Hutter:04book+}.

\subsection{A Computable Selector of Provably Optimal Actions}
\label{gm}

\Aixi above needs unlimited computation time. Its computable variant
\tl \cite{Hutter:04book+} has asymptotically optimal runtime but may suffer
from a huge constant slowdown.
To take the consumed computation time into account in a general,
optimal way, we may use the recent \gmn s 
\cite{Schmidhuber:05icann,Schmidhuber:05gmai,Schmidhuber:05gmconscious,Schmidhuber:09gm}
instead. They
represent the first class of mathematically rigorous, fully
self-referential, self-improving, general, optimally efficient problem solvers.
They are also applicable to the problem embodied by 
objective (\ref{optcur}).

The initial software  $\cal S$ of such a \gm contains
an initial problem solver, e.g., 
some typically sub-optimal  method
\cite{Kaelbling:96}.
It also contains an asymptotically optimal  initial proof
searcher based on an online variant of Levin's
{\em Universal Search} \cite{Levin:73},
which is used to run and test {\em proof techniques}. Proof techniques
are programs written in a universal language implemented
on the \gm within $\cal S$. They are in principle  able to compute proofs
concerning the system's own future performance, based on an axiomatic 
system $\cal A$ encoded in $\cal S$.
$\cal A$ describes the formal {\em utility} function, in our case eq. (\ref{optcur}),
the hardware properties, axioms of arithmetic and probability
theory and data manipulation etc, and  $\cal S$ itself, which is possible
without introducing circularity
\cite{Schmidhuber:09gm}.

Inspired by Kurt G\"{o}del's celebrated self-referential formulas (1931),
the \gm rewrites any part of its own code (including the proof searcher) through
a self-generated executable program as soon
as its {\em Universal Search} variant has found a proof that the rewrite is {\em useful} 
according to objective (\ref{optcur}).
According to the Global Optimality Theorem
\cite{Schmidhuber:05icann,Schmidhuber:05gmai,Schmidhuber:05gmconscious,Schmidhuber:09gm},
such a self-rewrite is globally optimal---no local maxima possible!---since
the self-referential code first had to prove that it is not useful to continue the 
search for alternative self-rewrites.

If there is no provably useful optimal
way of rewriting $\cal S$ at all, then humans
will not find one either. But if there is one,
then $\cal S$ itself can find and exploit it.  Unlike the previous {\em
non}-self-referential methods based on hardwired proof searchers \cite{Hutter:04book+},
\gmn s not only boast an optimal {\em order} of complexity but can optimally
reduce (through self-changes) any slowdowns hidden by the $O()$-notation, provided the utility
of such speed-ups is provable. Compare \cite{Schmidhuber:03newai,Schmidhuber:07newmillenniumai,Schmidhuber:06ai75}.

\subsection{Non-Universal But Still General and Practical RL Algorithms}
\label{rnn}
Recently there has been substantial progress in RL algorithms 
that are not quite as universal as those above, but nevertheless
capable of learning very general, program-like behavior.
In particular, evolutionary methods \cite{Rechenberg:71,Schwefel:74,Holland:75} 
can be used for training Recurrent Neural Networks (RNN),
which are general computers. Many approaches to
evolving RNN have been proposed
\cite{miller:icga89,yao:review93,yamauchi94sequential,nolfi:alife4,miglino95evolving,Sims:1994:EVC,Moriarty:98}.
One particularly effective  family of methods uses cooperative coevolution to
search the space  of network components ({\em neurons} or individual {\em synapses}) 
instead of complete
networks.  The components are {\em coevolved} by combining them into networks, and
selecting those for reproduction that participated in the best performing
networks~\cite{moriarty:ml96,gomez:ab97,gomez:ijcai99,gomez:phd,Gomez:03+,Gomez:08jmlr}.
Other recent RL 
techniques for RNN are based on the concept of 
 policy gradients \cite{Sutton:99,wierstraECML07,wierstraCEC08,rueckstiess2008b,sehnke2008b,wierstraPPSN08}.
It will be of interest to evaluate variants of
such control learning algorithms within
the curiosity reward framework.

\subsection{Acknowledgments}
\label{ack}
Thanks to Marcus Hutter, Andy Barto, Jonathan Lansey, Julian Togelius, Faustino J. Gomez,
Giovanni Pezzulo, Gianluca Baldassarre, Martin Butz,
for useful comments that helped to improve the first version of this paper.


\bibliography{bib}
\bibliographystyle{plain}
\end{document}